\documentclass{INTERSPEECH2023}

\usepackage{multirow}
\usepackage{caption}

\interspeechcameraready 

\title{Quantifying the perceptual value of lexical and non-lexical channels in speech}
\name{Sarenne Wallbridge, Peter Bell, Catherine Lai}

\address{
  Centre for Speech Technology Research, University of Edinburgh}
\email{\{s1301730, peter.bell, c.lai\}@ed.ac.uk}

\begin{document}

\maketitle
 
\begin{abstract}

Speech is a fundamental means of communication that can be seen to provide two channels for transmitting information: the lexical channel of \textit{which} words are said, and the non-lexical channel of \textit{how} they are spoken.
Both channels shape listener expectations of upcoming communication; however, directly quantifying their relative effect on expectations is challenging. Previous attempts require spoken variations of lexically-equivalent dialogue turns or conspicuous acoustic manipulations.
This paper introduces a generalised paradigm to study the value of non-lexical information in dialogue across unconstrained lexical content.
By quantifying the perceptual value of the non-lexical channel with both accuracy and entropy reduction, we show that non-lexical information produces a consistent effect on expectations of upcoming dialogue: even when it leads to poorer discriminative turn judgements than lexical content alone, it yields higher consensus among participants.

\end{abstract}

\section{Introduction}
\label{section:intro}
The human language system is often modelled as a predictive processor where expectations about the upcoming linguistic signal are conditioned on a host of contextual cues, including the previous signal \cite{Hale2001}.
These cognitive mechanisms have likely evolved to optimize our predictive capabilities for spoken interactions, a modality which can be framed as a multi-channel signal consisting of the lexical channel and non-lexical channels \cite{Christiansen2015, Wallbridge2021}. 
Although there is plentiful evidence that both channels are used to encode and decode information, the relative effect of these channels on human expectations in dialogue is unclear \cite{Silva2020, Pfau2000, Cohen1984}.  
In this work, we quantify the value of information in the lexical and non-lexical channels by how much it constrains human expectations regarding the upcoming 
dialogue turn. 
In particular, we address a previously-unanswered question of \textit{how the non-lexical channel affects expectations when the lexical channel is uninformative}. 


Attempts to disentangle channel 
effects often use acoustic manipulations such as low-pass filtering to delexicalise speech, and flattening pitch curves to remove changes in intonation \cite{Ruiter2006ProjectingTE, Bgels2015ListenersUI}. These modifications may be conspicuous and leave certain acoustic properties such as duration intact.  In previous work, we proposed the turn discrimination paradigm, which instead disentangles channel effects using separate lexical and acoustic conditions \cite{Wallbridge2021}. Rather than examining a specific function of non-lexical information, this method quantifies the value of non-lexical information as how much it affects performance of the generic task of discriminating between upcoming turns. 
By sampling contexts and responses from a conversational corpus, we 
showed that people use prosodic cues to discriminate the true response from alternative prosodic realisations in natural dialogue. This experimental design used sets of lexically-equivalent responses to isolate variation between responses to their prosodic realisations.
However, this severely limited our ability to quantify non-lexical channel value beyond short, often back-channel responses \cite{Ward2000ProsodicFW}, or across variable lexical content.

In this work, we present a generalisation of the turn discrimination paradigm using a dialogue-based language model (LM) to select lexically diverse but similarly plausible turns, as well as a novel quantification of channel value as entropy reduction to account for the inherent optionality in upcoming dialogue. 
This augmented paradigm enables investigation into the effect of non-lexical information on dialogue acceptability perception for  
variable lexical content, and allows us to more fully address the question of how non-lexical information is used when the lexical channel is uninformative.
We find that when the lexical channel is ambiguous, non-lexical information increases discriminative performance. However, when lexical content is informative, non-lexical information can worsen discrimination performance. Interestingly, it does so consistently: people tend to interpret non-lexical cues in similar ways even if this leads to incorrect judgements about the upcoming turn.

\section{Background}
\label{section:background}

\subsection{Speech: multi-channel communication}

As a communicative medium, speech encodes information in both lexical and non-lexical content.
Non-lexical information includes features of a speaker's identity and environment. However, in this work, we are primarily interested in prosodic information and how it affects dialogue perception. Generally quantified by the acoustic correlates for intonation, intensity, and rhythm (F0, energy, and unit duration),
prosody contributes to many important communicative functions such as marking novel information and topic shifts, conveying attitudinal reactions and uncertainty, and managing turn-taking \cite{Ward2019ProsodicPI}. 

There is experimental evidence for the interaction of lexical and prosodic information \cite{Gahl2004KnowledgeOG}; for example, \cite{Bgel2019FrequencyEA} shows that lexical and non-lexical channels are used jointly to mediate information density.
However, such studies often use carefully constructed stimuli 
or involve potentially conspicuous acoustic manipulation, a far cry from conversational speech which is rife with features like disfluencies and phonetic reduction \cite{Johnson2004MassiveRI}. 

Interest in the role of prosodic and lexical information in dialogue has a long history \cite{Black1997PredictingTI}.
Still, only a small number of psycholinguistic studies have explored the use of prosody in dialogue and for good reason \cite{Bgels2015ListenersUI, Tzeng2019CommunicativeCA, Lai2010WhatDY, FoxTree2000UntrainedSU}.  Disentangling lexical and non-lexical effects is unsurprisingly difficult given the complicated relationship between prosody and communicative intent, and the complexity of studying natural dialogue. 

Most similar to the work at hand is \cite{Bgels2021TurnendEI} which investigated the roles of prosody and conversational context for turn-end estimation. Using button-press experiments in conditions where participants either receive written transcripts or the full speech signal, phrase-final prosodic cues were found to be relevant for turn-end projection. However, their results reflect the complexity of processing and studying dialogue. Although acoustic context facilitated turn-end estimation for short utterances, it produced an inhibitory effect on the accuracy of turn-end estimation for longer turns.  Similarly, \cite{Corps2018PredictingTI} found that listeners leverage acoustic discourse context to make predictions. However, when acoustic context was relevant, participants' turn-end judgements were produced earlier, but not more precisely. 

\subsection{Optionality in communication}
While the studies above provide evidence that both the lexical and non-lexical channels affect dialogue perception, conclusions don't converge neatly. The majority of these studies \cite{Bgels2015ListenersUI, Tzeng2019CommunicativeCA, FoxTree2000UntrainedSU, Bgels2021TurnendEI, Corps2018PredictingTI} quantified channel value with accuracy-based metrics which don't account for the expected variability of the upcoming signal--its optionality. As text and speech generation approaches human-like naturalness, better understanding of appropriate production variability is increasingly important. 
We propose quantifying channel value in terms of entropy reduction to account for optionality in upcoming dialogue. 

The informational value of a message is often framed in terms of how surprising it is   
\cite{Shannon1948AMT, Hale2001, Levy2008}.
Highly surprising messages are informative but may be difficult to process. Conversely, if the future signal is perfectly predictable, there is little value in expending effort to communicate at all.
Recent work has demonstrated that, rather than being more probable, monological text that is perceived as natural encodes an amount of information that is close to the expected information content of natural language \cite{Meister2022HighPO}.
In other words, humans expect the upcoming linguistic signal to be within a certain interval of surprisal.
The degree of predictability in spoken dialogue is likely to be more complicated: it involves multiple parties with potentially different goals, and variation can occur across both the lexical and non-lexical channels. Enactment studies demonstrate the increased complexity of production variability in speech: different speakers produce different prosodic realisations of the same lexical content \cite{Mixdorff2012ProsodicS}.


\section{Experimental design}

\begin{figure}[t]
  \centering
  \includegraphics[scale=0.44]{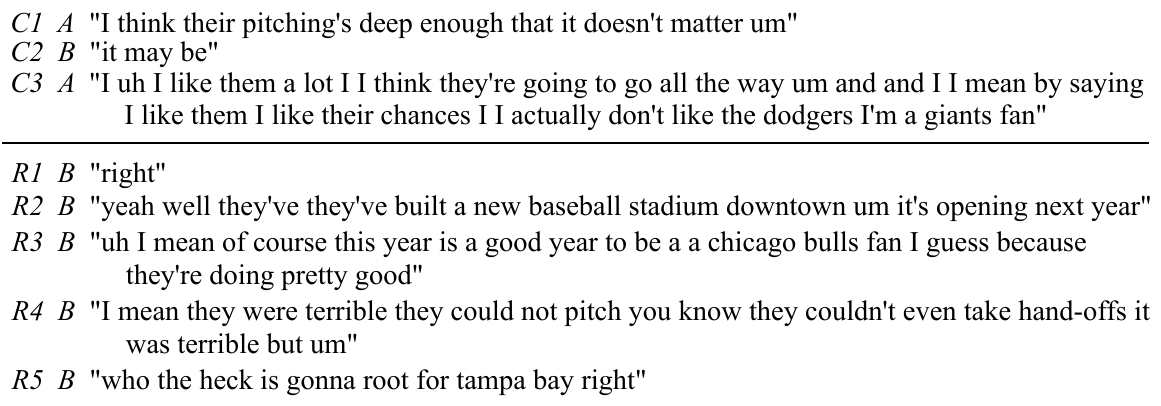}
  \caption{An example of plausible responses R[1:5] for speaker B sampled from our dialogue language model based on context turns C[1,2,3]}
  \label{fig:stimuli}
  \vspace{-16pt}
\end{figure}

\subsection{Generalised turn discrimination paradigm}
To quantify the perceptual value of lexical and non-lexical channels in constraining expectations of upcoming communication, we use the task of turn-acceptability presented in our previous work \cite{Wallbridge2021}: given conversational turns as context, participants are asked to rate the plausibility of potential continuations. Stimuli are presented as transcripts (lexical condition) or as speech recordings (acoustic condition). The value of non-lexical information is quantified by rating differences between conditions.

As described in the Introduction, the requirement of lexically-equivalent response sets in the original paradigm was a severe limitation. 
However, large LMs are known to align with aspects of human perception of monologues
\cite{Lau2017gradient, meister2021revisiting, wilcox2020predictive}.
Along with others \cite{Wallbridge2022, Giulianelli2021AnalysingHS}, we have recently shown that dialogue-based LM scores also correlate with human turn acceptability judgements \cite{Wallbridge2023}. Building on these recent findings, we remove the lexical-equivalence constraint by using a dialogue-based LM to sample plausible responses (model details are below). Figure \ref{fig:stimuli} displays a stimulus sampled from our model; responses are similarly plausible but lexically diverse.
This generalised methodology allows us to investigate the effect of non-lexical information on dialogue acceptability perception across unconstrained lexical content and variable responses.

\subsection{Task \& stimuli design}
Each stimuli consists of 3 contiguous turns of a conversation and five potential responses. Participants are instructed that one response is the true continuation for this conversation, then asked to score the plausibility of each response on a scale from 1 (`Very Unlikely') to 4 (`Very Likely'). We describe the stimuli construction and conditions below.

\textbf{Data} We construct stimuli from the Switchboard Telephone Corpus \cite{godfrey1992switchboard} which consists of over 2,400 dual-channel conversations between 542 speakers. 642 of these conversations were annotated post-hoc with information such as dialogue acts, information status, and prosodic features (Switchboard NXT \cite{Calhoun2010TheNS}). We use these conversations as validation and test sets. We segment all conversations into turns using the associated 
word timings.  First, all words spoken contiguously by a speaker are joined into segments. We remove completely overlapping segments 
before joining contiguous speaker segments into turns.

\textbf{Dialogue language model}
We use a state-of-the-art response selection model to sample plausible responses (cf. \cite{Wallbridge2023}). The architecture is a BERT cross-encoder post-trained and fine-tuned on Switchboard \cite{Han2021FinegrainedPF}.
Our model is implemented in PyTorch. We use \verb|bert-base-uncased| from the Transformers library as our base model and follow the training procedure from the original paper. Post-training augments the standard masked LM task with an utterance relevance classification task; 
response selection is the fine-tuning objective. Both train stages apply early stopping using our validation set of Switchboard. We post-train and fine-tune for 9 and 17 epochs, respectively.

\textbf{Transcript preprocessing} We clean the transcripts to train our dialogue-based LM and present stimuli to participants. In particular, characters specific to the Switchboard transcription guidelines\footnote{\url{https://isip.piconepress.com/projects/switchboard}} (mispronunciations, pronunciation variants, partial words, coinages) and non-speech sounds including vocalised noises and laughter are removed. 
We treat these as transcriptions of the non-lexical channel, rather than lexical content. 
However, the importance of filled pauses such as ``uh'' and ``um'' is widely accepted \cite{Clark2002UsingUA, Fraundorf2011TheDD}. We opt for the more conservative channel split and retain them in transcripts.

\textbf{Audio preprocessing} 
Overlapping speech is a prominent feature of conversation.
Although Switchboard recordings are dual-channel (one per speaker), certain conversations contain significant channel bleed. To maintain overlapping turns in our stimuli, we de-bleed all conversations: projections of the power spectra of both speaker channels are subtracted from each other before recomposing the individual channels into waveforms. 

\textbf{Stimuli construction} We randomly select a context as 3 contiguous speaker turns from the NXT conversations. We randomly sample 1000 unique responses to score with our dialogue LM. The five responses for this context consist of its true response, along with the top-scoring responses.
Random sampling allows us to explore the effect of non-lexical information across a range of linguistic functions. However, we perform broad filtering to ensure stimuli contain enough information but are not too long for the behavioural study. The final context turn must contain $3-50$ tokens with an audio length of $2-$\SI{10}{\second}. Potential responses must be of length $0.25-$\SI{10}{\second}, preceded by a pause within $-2 -$\SI{2}{\second}, and should not be from the same speaker as the context. The five first and last turns from all conversations are removed as the lexical content of greetings and farewells is highly conventionalised.
We construct 120 stimuli in total\footnote{We publish our stimuli: \url{https://sarenne.github.io/is-2023}}.

\textbf{Conditions} In both lexical and acoustic conditions, participants receive transcripts of the context. In the lexical condition, participants also receive transcripts of all potential responses; in the acoustic condition, responses are presented in audio format with each response spliced onto the final context turn.
Following \cite{Wallbridge2021}, pause length is treated as a feature of utterance design. Each response is thus extracted with its preceding pause which is used to join it with the final context turn.

\subsection{The online task \& stimuli sets}


Participants were recruited from Prolific Academic. We selected participants from North America for whom English was their first language to increase familiarity with the accents of speakers in Switchboard. Each participant received stimuli from one condition. Manually-constructed stimuli where only the true response was acceptable were interspersed throughout each survey as attention checks. Results of participants who obtained less than 80\% accuracy on these questions were removed (24\% and 26\% of participants in the lexical and acoustic conditions). Participants were presented with 20 random stimuli and the same five check questions. Average durations were $18.5 \pm 8$ (lexical) and $45.5 \pm 26$ (acoustic) minutes. In total, we collected 1200 responses: $5 \times 120$ stimuli in both conditions (45 and 50 participants in the lexical and acoustic conditions resp.)

\textbf{Stimuli sets}  
Our primary research question is whether participants use non-lexical information to guide expectations about an upcoming turn when the lexical channel is uninformative but diverse. To ensure that lexical content is uninformative, we create a subset of stimuli where participants did not reliably score the true response highest in the lexical condition (i.e., no more than two of the five participants ranked the true response as the only highest-scoring). From the 120 stimuli, this produced a subset of 63 ambiguous stimuli. 

\subsection{Metrics}

We use several metrics to quantify the perceptual value of acoustic information. Accuracy is easy to interpret, however, human acceptability judgements have been shown to be probabilistic \cite{Chater2006ProbabilisticMO, Lau2017gradient, Wallbridge2023}. As such, it is possible for multiple responses to be considered plausible. To account for this optionality, we employ entropy reduction to examine the convergence of participants' judgements at both response- and question-levels. 
Entropy has previously been used to quantify the potential value of non-lexical component of spoken dialogue \cite{Ward2009EstimatingTP}.

\textbf{Accuracy} Accuracy reflects the frequency with which the true response was rated highest. We also weight this frequency by the proportion of score mass assigned to the true response by each participant (weighted accuracy).

\textbf{Ordinal Entropy} We measure entropy-per-response using a variant of cumulative paired $\phi$-entropy to account for the ordinal nature of scores \cite{Klein2016CumulativeP}. Standard entropy $H(S)$ is a function of categorical label probabilities. Ordinal entropy $H_{Ord}(S)$ is instead a function of the cumulative probability for each score, thus reflecting dispersion among scores \cite{Yager2001DissonanceAM}.

\textbf{Permutation Entropy} Permutation entropy measures entropy at the stimulus-level. Ordinal Pattern Analysis (OPA) is often used to quantify the complexity of time-series data by converting it to a sequence of ordinal patterns before computing standard Shannon entropy $H(S)$ across pattern frequencies \cite{cuesta2019permutation}. We convert participant scores across responses to rank patterns. This quantifies agreement at the stimulus-level.


\vspace{-4mm}
\begin{gather}
    H(S) = - \sum_{i=1}^n{p_{i} \log p_{i}} \\
    H_{ord}(S) = - \sum_{i=1}^n{\Big( p_{\le k} \log p_{\le k} -  (1 - p_{\le k}) \log (1-p_{\le k}) \Big)}
\end{gather}  
\vspace{-3mm}

For score counts $S_i$ over scores $i \in 1,...,n$, we denote $p_i = P(S_i)$ and $p_{\le k} = \sum_{i=1}^k p_i$

\section{Results \& Discussion}

To test whether participants use non-lexical information to guide their expectations when the lexical channel is uninformative but unconstrained, we compare scoring behaviour in the lexical and acoustic conditions across the ambiguous stimuli.  Chance performance is estimated by shuffling each set of participant ratings 100 times to maintain score distributions.

Additionally, we analyze the effect of condition on ordinal entropy
while controlling for other factors with Bayesian multilevel regression models. As entropy values are bounded and continuous, we scale them to $[0,1]$ and use Zero-One Inflated Beta Regression.
Models were fit using \verb|brms| in R \cite{brms}. 
To investigate potential cognitive load differences, we include \emph{response length} (in seconds) as a predictor.  
Following \cite{Wallbridge2022}, we include the \emph{mean surprisal} of the response conditioned on the context, and an indicator for whether the response was the \emph{true} continuation to see if participants treated true and false continuations differently. 
Group-level effects (i.e., random effects) for context and response dialogue acts from the Switchboard NXT annotations \cite{Calhoun2010TheNS}, as well as effects for context, and context-response identifiers are included to control for stimuli variation.
We include interaction terms with response length, target, mean surprisal, and dialogue acts to see if effects varied with the acoustic/lexical condition.  We see non-zero variance estimates for the group-level effects, i.e., these factors do account for variation in the entropy. 

We use the emmeans package \cite{emmeans} to compute estimated marginal means and 95\% Highest Posterior Density Regions, i.e. Credible Intervals (CIs), to examine effects of predictors.

\begin{table}[t]
  \centering
  \caption{Evaluations of ambiguous stimuli across conditions, and the mean per-question difference between them (standard, weighted accuracy ($acc$, $acc_{\{W\}}$), ordinal, permutation entropy ($H_{Ord}$, $H_{Perm}$); all normalised). Differences are all significant using a directional Wilcoxon rank sum test ($p<0.002$).}
  \label{tab:ambiguous_eval}
  \vspace{-5pt}
    \begin{tabular}{lcccc}
        \hline
         &  \multicolumn{4}{c}{\textbf{Condition}} \\ 
        \cmidrule{2-5}
        \textbf{Metric} & \textbf{\textit{Chance}} & \textbf{Lexical} & \textbf{Acoustic} & \textbf{Difference} \\
        \midrule
        $acc$         & $0.07 \pm 0.01$ & 0.08 & 0.25 & 0.16 \\
        $acc_{\{W\}}$ & $0.05 \pm 0.01$ & 0.05 & 0.17 & 0.11 \\ 
        $H_{Ord}$     & $0.66 \pm 0.01$ & 0.54 & 0.48 & -0.06 \\
        $H_{Perm}$    & $0.94 \pm 0.01$ & 0.90 & 0.86 & -0.04 \\
    \hline
    \end{tabular}
  \vspace{-6pt}
\end{table}

\begin{table}[t]
  \centering
  \caption{Accuracy metrics for check questions}
  \label{tab:check_eval}
  \vspace{-5pt}
    \begin{tabular}{lccc}
        \hline
         &  \multicolumn{3}{c}{\textbf{Condition}} \\ 
        \cmidrule{2-4}
        \textbf{Metric} & \textbf{Lexical} & \textbf{Acoustic} & \textbf{Difference} \\
        \midrule
        $acc$         & 0.96 & 0.96 & 0.00 \\
        $acc_{\{W\}}$ & 0.91 & 0.74 & -0.17 \\ 
    \hline
    \end{tabular}
  \vspace{-15pt}
\end{table}

\begin{table}[t]
  \centering
  \caption{Evaluations of remaining stimuli. Mean per-question differences are all significant using a directional Wilcoxon rank sum test at $p<0.0001$ except permutation entropy ($p<0.02$).}
  \label{tab:nonambig_eval}
  \vspace{-5pt}
    \begin{tabular}{lcccc}
        \hline
         &  \multicolumn{4}{c}{\textbf{Condition}} \\  
        \cmidrule{2-5}
        \textbf{Metric} & \textbf{\textit{Chance}} & \textbf{Lexical} & \textbf{Acoustic} & \textbf{Difference} \\
        \midrule
        $acc$         & $0.14 \pm 0.02$ & 0.54 & 0.36 & -0.18 \\
        $acc_{\{W\}}$ & $0.10 \pm 0.01$ & 0.38 & 0.25 & -0.12 \\ 
        $H_{Ord}$     & $0.70 \pm 0.01$ & 0.55 & 0.48 & -0.06 \\
        $H_{Perm}$    & $0.96 \pm 0.01$ & 0.89 & 0.86 & -0.03 \\
    \hline
    \end{tabular}
  \vspace{-15pt}
\end{table}

\subsection{Accuracy}
As can be seen in Table \ref{tab:ambiguous_eval}, accuracy for \emph{ambiguous} stimuli in the lexical condition is close to chance. In the acoustic condition, it is significantly higher.
Increased accuracy provides strong evidence that participants leverage non-lexical cues to constrain their expectations about the upcoming dialogue turn. 

Table \ref{tab:nonambig_eval} contains results for the \emph{remaining} stimuli--where participants could discriminate the true response relatively accurately from the lexical channel alone. People judge these stimuli more accurately. However, their accuracy drops in the acoustic condition. Surprisingly, people are less apt at selecting the true turn when provided with non-lexical information.




Accuracy metrics for the check questions are shown in Table \ref{tab:check_eval}.  Weighted accuracy decreases by $-0.17$ in the acoustic condition, reflecting less decisive scores for check questions.  This is surprising as multi-modal communication is suggested to offer greater communicative power than single-channel communication
\cite{Frhlich2019MultimodalCA,Hebets2004ComplexSF}. 
We hypothesise that acoustic stimuli place a higher cognitive load on participants. As such, these results likely understate non-lexical channel value. 

\subsection{Entropy Reduction}
Given that the upcoming communicative signal is not perfectly predictable, we also quantify the effect of non-lexical information as entropy reduction over participant scores. 
Both entropy-based metrics decrease in the acoustic condition for the \emph{ambiguous} 
stimuli--participants agree more in the acoustic condition (Table \ref{tab:ambiguous_eval}). This suggests that the non-lexical channel provides additional cues for what may come next and that participants interpret them similarly. Crucially, although Table \ref{tab:nonambig_eval} shows reduced accuracy between conditions for the \emph{remaining} stimuli, score entropies decrease and to similar degrees as found across the lexically-ambiguous stimuli.
The non-lexical channel seems to produce consistent perceptual effects, regardless of how informative lexical content is for turn discrimination.



Higher entropy in the lexical condition is reflected by our regression model, but the difference varies depending on other factors. In particular, we see an interaction between condition and response length: a positive slope estimate in the acoustic condition (0.02, CI=(0.01, 0.03)), and a flat slope for the lexical condition ($-0.001$, CI=($-0.01$, 0.01)). 
That is, participants show higher agreement in the acoustic condition than the lexical condition for short responses, but the difference shrinks with utterance length. This further suggests potential differences in cognitive load between conditions.

We calculate estimated marginal mean ordinal entropy for true and false responses (by condition), while averaging over other predictors, to see their effect. 
We see lower entropy for true responses overall.  Similarly, entropy is reduced in the acoustic condition compared to the lexical condition. However, the difference between conditions is greater for true responses ($-0.13$, CI=($-0.26$,$-0.001$) vs. ($-0.10$, CI=($-0.20$, 0.001)). This suggests that true responses have acoustic features that participants make use of for this task.

Interestingly, surprisal affects ordinal entropy differently between conditions. The estimated effect is likely positive in the acoustic condition (0.08, CI=($-0.001$, 0.16)), i.e., more lexically surprising responses result in less agreement. However, the effect in the lexical condition peaks around zero (0.00,  CI=($-0.07$, 0.08)). The latter estimate is likely a result of the ambiguous stimuli selection, but also again indicates that acoustic information changes participant expectations.

\section{Conclusions}

The generalised turn-discrimination paradigm presented here enabled our analysis of how lexical and non-lexical channels are used jointly for a much broader set of language than was previously possible. Our results provide firm evidence that non-lexical information constrains expectations of spoken dialogue--people can leverage non-lexical cues to discriminate true continuations from false candidates when the lexical channel is uninformative.
However, when the lexical content is informative for the discriminative task, acoustic information can hinder performance. 
Although surprising, similar results were found by \cite{Corps2018PredictingTI} who showed that listeners respond earlier but not necessarily more accurately in turn-end detection tasks when context is informative. 
Quantifying channel value as entropy reduction provides a novel perspective on channel value: even when it leads to incorrect discriminative judgements, the non-lexical channel affects expectations in consistent ways.

We believe our methodology has implications both for learning perceptually-motivated representations of spoken communication and for speech generation where the degree of acceptable production variability across both lexical and non-lexical channels in dialogue is relatively unexplored. Here, we investigated a small number of factors that could affect non-lexical channel value and found evidence of complex interactions with acoustics; in future work, we hope to develop a more formal conditional quantification of channel value. 
For example, different speech acts have been shown to exhibit greater prosodic variation, potentially indicating that the information mass is skewed more heavily towards the non-lexical channel for certain speech acts \cite{Syrdal08}.


\textbf{Acknowledgements}
We'd like to thank Erfan Loweimi and Cassia Valentini Botinhao for help with channel bleed removal.





\bibliographystyle{IEEEtran}
\bibliography{mybib}

\end{document}